\documentclass{Interspeech2024}

\usepackage{pifont}
\usepackage{multirow}
\usepackage{float}
\usepackage{cleveref}

\usepackage{cjhebrew}

\newcommand{\newpara}[1]{\vspace{0.2cm}\noindent \textbf{#1}}

\interspeechcameraready

\title{A Language Modeling Approach to Diacritic-Free Hebrew TTS}

\name[]{Amit}{Roth}
\name[]{Arnon}{Turetzky}
\name[]{Yossi}{Adi}

\address{School of Computer Science and Engineering\\
The Hebrew University of Jerusalem, Israel}
\email{amit.roth@mail.huji.ac.il}
\keywords{Text-to-Speech, Diacritic, Hebrew speech}

\begin{document}

\maketitle
\begin{abstract}
We tackle the task of text-to-speech (TTS) in Hebrew. Traditional Hebrew contains \emph{Diacritics}, which dictate the way individuals should pronounce given words, however, modern Hebrew rarely uses them. The lack of diacritics in modern Hebrew results in readers expected to conclude the correct pronunciation and understand which phonemes to use based on the context. This imposes a fundamental challenge on TTS systems to accurately map between text-to-speech. In this work, we propose to adopt a language modeling Diacritics-Free approach, for the task of Hebrew TTS. The model operates on discrete speech representations and is conditioned on a word-piece tokenizer. We optimize the proposed method using in-the-wild weakly supervised data and compare it to several diacritic-based TTS systems. Results suggest the proposed method is superior to the evaluated baselines considering both content preservation and naturalness of the generated speech. Samples can be found under the following link: \url{pages.cs.huji.ac.il/adiyoss-lab/HebTTS/}
\end{abstract}

\section{Introduction}
\label{sec:intro}
Hebrew, a low-resource language spoken by $9$ million people worldwide~\cite{campbell2008ethnologue}, presents unique challenges that constrain research and product development in speech technology. Specifically, Hebrew is a morphologically rich language, with the common use of prefixes and suffixes to modify words’ meanings and to add prepositions. On top of that, Hebrew uses Diacritics ('Niqqud') to create a one-to-one mapping between text and phonemes. 'Niqqud' is a system of Diacritical signs used to represent vowels or distinguish between alternative pronunciations of letters of the Hebrew alphabet. 

In practice, modern Hebrew text rarely contains Diacritics, one may find Diacriticized text in specialized texts such as dictionaries, poetry, or children's books. Hence, readers are expected to conclude the correct pronunciation and understand which phonemes to use, based on familiarity with the language itself. This makes it challenging for text-to-speech (TTS) systems to accurately learn the connection between text and speech. For example the words: \begin{cjhebrew}\cjRL{mat*AnAh}\end{cjhebrew} and \begin{cjhebrew}\cjRL{mat:nEh}\end{cjhebrew}, contain the same characters, but have completely different meaning and pronunciation, the first one means a `gift', while the second one means `conditioning'. As mentioned before, in modern Hebrew writing, one will probably not encounter Diacritics and the above word will appear as follows, \begin{cjhebrew}\cjRL{mtnh}\end{cjhebrew}. As a result, the reader should infer the right pronunciation by context only. Moreover, when considering spoken language modeling systems in automated pipelines, current Automatic Speech Recognition (ASR) systems, such as Whisper~\cite{radford2022robust} and Massively Multilingual Speech (MMS)~\cite{pratap2023scaling} do not output diacritics in their transcripts, hence TTS systems should either predict it or use a Diacritic-free synthesis system. 

\begin{figure}[t!]
  \centering
  \includegraphics[width=0.9\linewidth]{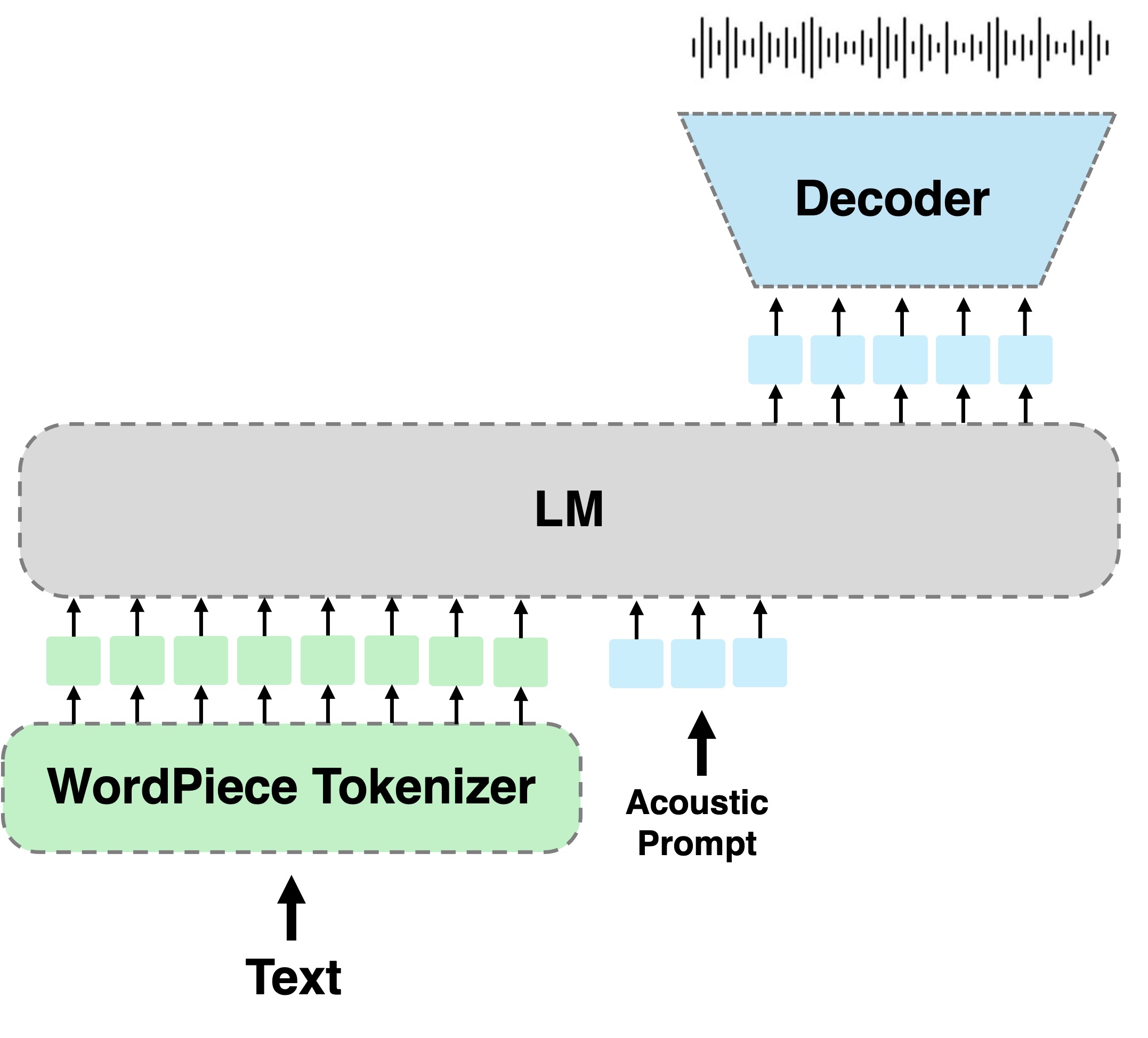}
  \caption{A high-level overview of the the proposed method. The text is first being tokenized using a word-piece tokenizer. Then an audio language model predicts a discrete sequence of audio tokens which later on will be decoded into raw waveform. \vspace{-0.2cm}}
  \label{fig:model}
\end{figure}

Another major issue that holds progress in developing AI-based Hebrew TTS systems is the lack of datasets. As Hebrew is considered a low-resource language, public spoken benchmarks hardly exist. Previous efforts in constructing datasets in Hebrew were often relatively small~\cite{izre2001designing, azogui2016open, marmorstein2022huji, sharoni2023saspeech}. The authors in~\cite{izre2001designing} established the Corpus of Spoken Israeli Hebrew (CoSIH) with the goal of compiling a large database of recordings of spoken Israeli Hebrew in order to facilitate and enhance research in the field. Next, the authors in~\cite{azogui2016open} released The Map Task Corpus (MaTaCOp) of Hebrew dialogues. The authors in ~\cite{marmorstein2022huji} collected naturally occurring speech and interaction in Modern Hebrew via telephone conversations during the years 2020–2021 and released the HUJI Corpus of Spoken Hebrew (HUJICorpus). More recently, the authors in ~\cite{sharoni2023saspeech} released SASPEECH, a high-quality single-speaker Hebrew dataset to enhance Hebrew speech synthesis research. Although all of these prior work are important and valuable, the provided benchmarks are relatively small. CoSIH contains $\sim12.3$ hours of speech, the MaTaCOp corpus contains $\sim5.3$ hours, the HUJI Corpus has $\sim3.8$, and SASPEECH which is the largest one contains $\sim30$ hours of speech. For comparison a modern, contextualized TTS system in English is trained over $\sim60$k hours~\cite{wang2023neural}. Recently, the authors of~\cite{marmor2023ivrit} and the authors of ~\cite{turetzky2024hebdb} released two datasets denoted as \emph{ivrit.ai} and \emph{HebDB} respectively. The authors released weakly-supervised speech from local podcasts and provided the first large-scale dataset in Hebrew, which we leveraged to construct the model. 

Previous attempts were made to construct a TTS system in Hebrew. The Authors of~\cite{pratap2023scaling}, proposed the MMS system. In their study, they develop speech technologies (ASR, TTS, Language ID) in more than $1,000$ languages. Their TTS system is based on representation obtained from a pre-trained multi-lingual self-supervised model. Although providing impressive results, their Hebrew TTS system is based on predicting diacritics of the input text. More recently, the authors of~\cite{sharoni2023saspeech} introduced the Overflow~\cite{Mehta_2023} model for Hebrew, together with the SPASEECH benchmark. The Overflow model is comprised of neural HMM together with normalizing flows. On top of the Overflow model, the authors in~\cite{sharoni2023saspeech} suggested using the HiFi-GAN neural vocoder~\cite{kong2020hifigan} to estimate the phase. Similarly to MMS, the system proposed by~\cite{sharoni2023saspeech} is based on predicting diacritics of the input text, hence is sub-optimal and often produces wrong and unnatural content in the generated speech. Moreover, such dependency makes it difficult for these models to scale to large datasets as they both require predicting diacritics on top of automatically transcribed text. Unlike these methods, the proposed LM approach operates in a Diacritic-free manner, not propagating mistakes from the diacritic prediction models, and better leveraging the context of the input signal.

Recent studies in speech and audio representation learning proposed learning discrete representation of the input signal~\cite{défossez2022high, zeghidour2021soundstream}. Such representation can be later used for plenty of speech and audio synthesis tasks~\cite{copet2024simple, kreuk2023audiogen, borsos2023audiolm, sheffer2023hear}. Specifically, the authors of~\cite{wang2023neural, kharitonov2023speak, lyth2024natural} proposed optimizing an LM on top of such discrete speech representation, conditioned on a phonemic representation of the input text for the task of TTS. Following such an approach was found to produce high-quality and natural speech, with the ability to rapidly adapt to new speakers via acoustic prompting. As this approach is contextualized by nature it may serve as the ideal candidate for a Diacritic-free Hebrew TTS system. 

In this work, we study and propose a \emph{Language Modeling} approach which operates over discrete representations of the speech signal to construct a Hebrew TTS system. We optimize an acoustic LM over a weakly supervised large-scale dataset containing in-the-wild recordings. We empirically demonstrate that following the LM approach makes the usage of diacritics in Hebrew redundant, hence yielding a diacritic-free approach. We study several text representation methods and found that using word-piece tokenization produces the best results overall. Results suggest the proposed method is superior to the evaluated baselines considering both content preservation and generation quality. Code, dataset, and models are publicly available under the following link: \url{https://pages.cs.huji.ac.il/adiyoss-lab/HebTTS/}.
\section{Method}
\label{sec:method}

Given a dataset $D=\{x_i, y_i\}$ where $y_i$ is an audio sample and $x_i$ is its corresponding transcription. We encode the audio into a sequence of discrete units. We follow the approach proposed by~\cite{défossez2022high} and encode the audio using Residual Vector Quantization (RVQ). Formally, $E(y) = C^{T\times N_{cb}}$ where $C$ represents the acoustic code matrix, where $N_{cb}$ is the number of codebooks and $T$ is the utterance length. 

A common paradigm in the TTS field is to represent text in its most basic form, i.e., phonemes~\cite{tan2021survey}. As we aim at building a Diacritic-free system we can not use phonemes as text representations. As a result, we use a sub-word tokenizer in the form of word-piece tokenization. Such tokenizer was found beneficial in text encoders such as BERT~\cite{devlin2019bert}, and more relevant to our setup AlephBERT~\cite{alephBert2021}. We experimented with several other tokenizers, however, found the word-piece to provide the best overall results (see Section~\ref{sec:toke} for more details). Below we describe both text tokenization and model in more detail. We depict a general overview of the LM-based approach in Fig.~\ref{fig:model}. 

\begin{table}[t!]
  \caption{A comparison between prior, non-LM-based TTS systems against the proposed system. Prior work is mainly based on Mel-spectrogram, Diacritics, and relatively small amounts of training data. We show that while following the LM approach we can leverage large amounts of in-the-wild training data, using plain text, on top of discrete learned speech representations.\label{tab:sys}}
  \centering
  \resizebox{0.48\textwidth}{!}{
  \begin{tabular}{l|ll}
    \toprule
    & {\textbf{Prior work}} & {\textbf{Proposed sys.}}\\
    \midrule
     Intermediate Rep.           &   Mel spectrogram  & Audio discrete rep.\\
     Training data (type)        &   Niqqud            & Plain text \\
     Training data (hours)       &   $\sim$30 hours   & $\sim$5.0k hours   \\
    \bottomrule
  \end{tabular}}
  \vspace{-0.2cm}
\end{table}

\newpara{Text Tokens.}
\label{sec:text_tok}
We tokenize the text using a word-piece text tokenizer similar to the one proposed by~\cite{alephBert2021}. Specifically, we leverage a pre-trained Hebrew text tokenizer that was trained using $98.7$M Hebrew sentences.
word-piece tokenizers were tested in different models~\cite{devlin2019bert, wu2016google, schuster2012japanese} and performs similarly to Byte-Pair Encoding~\cite{Gage1994ANA}.

Given a training corpus $C$ and a number of desired word-pieces $t$, the optimization problem is to select $t$ word-pieces such that minimizes the number of word-pieces generated when tokenizing the entire corpus $C$.
We start with a small character vocabulary and special tokens $W$, and apply merge rules for the elements. iteratively we compute for each pair $w_1, w_2 \in W$ a score as seen in equation~\ref{eq:wordpiece} and merge the pair with the maximum score getting a new vocabulary $W' = W \cup \{(w_1, w_2)\}$. We follow this step with the new vocabulary until $|W| = t$.

\begin{equation}
    score = \frac{freq(e_1, e_2)}{freq(e_1) \times freq (e_2)},
    \label{eq:wordpiece}
\end{equation}
By dividing the frequency of the pair by the product of the frequencies of each of its parts, the algorithm prioritizes the merging of pairs where the individual parts are less frequent in the vocabulary. 

\begin{table*}[t!]
  \caption{Comparison of the LM based approach to both MMS~\cite{pratap2023scaling} and Overflow~\cite{sharoni2023saspeech}. We report both objective metrics (WER, CER, and speaker similarity), together with two human studies evaluating the naturalness and content preservation in the generated samples. In the human study, we report mean and standard deviations.\label{tab:mainres}}  
  \centering
  \begin{tabular}{l|ccc|cc}
    \toprule
    & \multicolumn{3}{c|}{\bf Objective Metrics} & \multicolumn{2}{c}{\bf Human Study} \\
    \midrule
    \textbf{Model} & \textbf{WER} & \textbf{CER} & \bf Spk. Sim. & \textbf{Naturalness} & \textbf{Content}\\
    \midrule
    Reference & 0.07 & 0.03 & 0.97 & 4.68 ($\pm 0.46$) & 4.63 ($\pm 0.51$)\\
    \midrule
    MMS~\cite{pratap2023scaling}            & 0.23   & 0.07 & - & 2.51 ($\pm 1.05$) &2.35 ($\pm 0.77$)  \\
    Overflow~\cite{sharoni2023saspeech}     & 0.20   & 0.08 & 0.88 & 3.44 ($\pm 1.01$) & 3.79 ($\pm 0.77$)\\
    Ours (seen speaker)                     & 0.19 & 0.08   &\textbf{0.95}& \textbf{4.17 ($\pm 0.80$)} &4.44 ($\pm 0.68$)\\
    Ours (unseen speaker)                   & 0.19 & 0.08  &0.92 & 4.05 ($\pm 0.75$) & \textbf{4.48 ($\pm 0.58$)}\\
    \bottomrule
  \end{tabular}
\end{table*}

\newpara{Model.}
Recall, our goal is to produce a Diacritic-free Hebrew TTS system that can handle weakly transcribed spoken data. Hence, we proposed leveraging the abilities of language models to efficiently model long contexts. Inspired by recent LM-based approaches for TTS~\cite{wang2023neural, kharitonov2023speak}, our model uses an LM approach that operates directly over discrete representation obtained from a pre-trained speech encoder.  

The model first receives a text prompt, $x_i$, and a $3$-second enrolled recording as an acoustic prompt. We then, encode the acoustic prompt using the same speech encoder $E$, and process the text using the text tokenizer defined in sub-section~\ref{sec:text_tok}. Recall, that the speech encoder, $E$, quantizes the utterance using RVQ module, hence it outputs a matrix of size $T\times N_{cb}$. Meaning that, at each time step we are left with $N_{cb}$ discrete codes. 

There are several alternatives in the literature to handle this complex input structure. For instance, the authors in~\cite{copet2024simple, kharitonov2022text} proposed to predict all codes at each time-step in parallel while introducing a delay pattern to better model the conditional probability distribution. The authors in~\cite{borsos2023audiolm} proposed flattening the whole sequence (resulting in a $N_{cb}$ times larger sequence) and splitting its modeling across two LMs. 

In this work, we follow the approach proposed in~\cite{wang2023neural}. In which, the first codebook, $c_{,:1}$, is modeled in an Auto-Regressive (AR) manner following the standard next token prediction framework. Specifically, we concatenate the word-piece tokens with the first codebook from the acoustic prompt, denoted by $w, c_{\le t,1}$, to infer the next acoustic token $c_{t,1}$ of the target signal. The rest of the codebooks (2 to $N_{cb}$), are modeled using a non-autoregressive (NAR) model, where the network is trained to maximize the acoustic tokens likelihood derived from the $i$-th quantizer codebook, conditioned on the sum of all representations from previous codebooks.

Overall, unlike, prior works which are mainly based on Mel-spectrogram as speech representations, diacritics for text, and relatively small and high-quality amounts of training data. Following the LM approach, allows us to leverage large amounts of in-the-wild recordings, using plain text, and operate on top of discrete learned speech representations. Table~\ref{tab:sys} summarizes the main differences between the methods. 
\section{Dataset}
\label{sec:data}
We use both the \emph{ivrit.ai} dataset~\cite{marmor2023ivrit} together the HebDB dataset~\cite{turetzky2024hebdb}. Both datasets consists of $\sim4500$ hours of speech gathered from local podcasts ($\sim1700$ from HebDB and $\sim2800$ from \emph{ivrit.ai}). These datasets are comprised of spontaneous dialogues, featuring multiple speakers discussing a wide range of topics including economy, politics, sports, culture, science, history, and music, among others. The podcast recordings are full episodes, thus containing lengthy audio tracks and various non-speech elements such as music, environmental sounds, and periods of silence. Such real-world conditions present challenges for model optimization and necessitate preprocessing steps. We apply the same pre-processing pipeline to both \emph{ivrit.ai} dataset to all the dataset. Initially, we standardize all audio recordings to a consistent $16$kHz, mono recordings, using julius~\footnote{\url{https://github.com/adefossez/julius}} python package.  Subsequently, we employ a Voice Activity Detection (VAD) model, namely \texttt{silero-vad}~\cite{silero_vad} to perform a voice activity detection and segment the waveforms into sentences, filtering out activated segments with a minimum duration of $1$ seconds, separating audio segments by a minimal silence duration of $100$ms and padding both sides of the segmented audio with $30$ms of silence. Finally, we automatically transcribe the segmented speech using a pre-trained ASR model, specifically Whisper V2-Large \cite{radford2022robust}. After preprocessing our data, we are left with $\sim4500$ hours of natural dialogues with weakly labeled transcriptions.
\section{Experiment Setup}
\label{sec:setup}
\subsection{Implementation details}
Our model contains $420$M parameters and is trained on $8$ NVIDIA A$30$ $24$GB GPUs with a total batch size of $144,000$ acoustic tokens. We optimize the model using \emph{EDEN} scheduler as used in~\cite{yao2024zipformer} with a starting learning rate of $5 \times 10^{-2}$. We train the AR model for $1.2$M steps and the NAR for $200$k steps. 
For the audio tokenizer, we use the officially released pretrained version of EnCodec~\cite{défossez2022high} sampled at $24$Khz to generate acoustic tokens~\footnote{\url{https://github.com/facebookresearch/audiocraft/blob/main/docs/ENCODEC.md}}. To improve the quality of the generated audio we use the pre-trained Multi Band Diffusion (MBD) vocoder \cite{sanroman2023fromdi}. For tokenization, we use the pretrained word-piece tokenizer of AlephBERT~\footnote{\url{https://github.com/OnlpLab/AlephBERT}} with vocabulary size of $52$k tokens. We train the model for audio length sequences between $1-18$ seconds. We sample the 50 most likely tokens using $top_k=50$ and temperature = $1$.
We adopt the following public code~\footnote{\url{https://github.com/lifeiteng/vall-e}}. 

\subsection{Evaluation metrics}
We evaluate the proposed method considering both objective metrics and human study. We consider several axes: (i) content preservation in the form of Word Error Rate (WER), Character Error Rate (CER), and human study; (ii) speaker similarity using a pre-trained speaker verification model; and (iii) overall quality and naturalness via human study. We describe each of the evaluation metrics below.  

\newpara{WER and CER.} We calculated Word Error Rates (WERs) and Character Error Rates (CERs) between the input text and an automatic transcription generated by an ASR system. Specifically, we run this evaluation using $100$ randomly sampled text prompts with diacritics from SASPEECH \cite{sharoni2023saspeech} dataset. We remove the diacritics for our model and compare with the transcribed text from Whisper V2-Large \cite{radford2022robust} model which provides state-of-the-art performance. We normalize the text by removing all punctuation from both original and transcribed text. To improve the robustness of the sampling process, we sample three audio generations for each input prompt and select the one with the best WER w.r.t the input text. To calibrate the results with the errors produced by the Whisper model, we additionally calculate WER and CER between the reference and transcribed text of the original recordings.

\newpara{Speaker similarity.} For speaker similarity we measure the cosine similarity between the generated speaker and an enrollment set of five different recordings of the person to identify. To compute the cosine similarity we use a state-of-the-art pre-trained speaker verification model~\cite{Chen_2022}. This similarity measure was found to be beneficial in prior work~\cite{polyak2021speech, wang2023neural, wang2021fairseq}. 

\newpara{Human evaluation.} We conduct two different human studies to evaluate the quality of the generated samples. Raters were asked to evaluate the quality of the generated speech considering both generation fidelity and naturalness of the speech signals on a scale between 1 -- 5, where 1 has the lowest quality and 5 is the best quality. We evaluate $20$ samples from each of the evaluated methods while we enforce at least $15$ ratings for each sample. All raters are native Hebrew speakers. 

Although the Whisper model reaches state-of-the-art performance, its WER in Hebrew is still $\sim27$. Hence, we additionally ask raters to rate the accuracy between the generated speech and the written text. Same as before raters evaluated the content of the recordings on a scale of 1 -- 5, where 1 is the least accurate and 5 has a perfect match. We conduct a human study to evaluate the proposed method against the baseline methods as well as to evaluate the text tokenization method. 

\subsection{Baseline systems}
We compare the proposed method against two baseline systems: (i) Massively Multilingual Speech (MMS)~\cite{pratap2023scaling} and Overflow~\cite{sharoni2023saspeech}. The MMS model is based on a multi-lingual wav2vec2.0~\cite{baevski2020wav2vec} trained on $\sim500$k hours from $1,107$ languages, while $25$ hours in Hebrew. The Overflow model is based on a neural HMM combined with normalizing flows for describing highly non-Gaussian distribution of the acoustics. This model was trained over $30$ hours of single-speaker, high-quality data, obtained from the `Hayot-Kiss' podcast~\cite{sharoni2023saspeech}. 
Both methods are based on predicting Diacritics using an external model. In both methods, we use the official pre-trained models released by the authors and follow exactly their text pre-processing pipelines. 
\section{Results}
\label{sec:res}

We start by evaluating the proposed method against both MMS and Overflow. Results are summarized in Table~\ref{tab:mainres}. The proposed method provides superior performance to the evaluated baselines considering both objective metrics and human study. Notice, following the LM approach for Hebrew TTS additionally, allows fast adaptation to new speakers. The proposed method shows minor differences in performance when considering speech and unseen speakers. 

Interestingly, when considering WER, CER, and Speaker similarity, the Overflow method provides comparable performance to ours while being superior to the MMS model. The main difference between the methods is reflected in the naturalness of the generated speech. Moreover, it is worth mentioning that although the WER and CER are comparable across all methods (with MMS achieving worse WER and Overflow achieving worse CER), these are based on automatic transcriptions that do not take into account the pronunciation, meaning two different words can be transcribed to the same sequence characters while reflecting completely different pronunciation. However, when investigating the content metric under the human study we observe larger differences. 

\newpara{The effect of the tokenizer.}
\label{sec:toke}
As there is no direct mapping between non-diacritic text to phonemes in Hebrew, it is not clear how one should represent the text for the system. A natural approach would be to use character tokenization (i.e., converting the text into a sequence of characters). Another alternative that gains popularity in textual language models is to use a word-piece tokenizer~\cite{minaee2024large}. In this study, we follow the word-piece tokenizer approach. 

To better evaluate the effect of using different tokenization methods for the input text we trained two versions of the proposed method using both chars and word-piece tokenizer. We additionally experimented with contextualized representations obtained from hidden layers of a pre-trained text encoder model, namely AlephBERT~\cite{alephBert2021}. Unfortunately, such text representation performs significantly worse than the other tokenizers, hence we do not report results for it. We measure WER and CER metrics, together with a human study measuring content preservation. Results are presented in Table~\ref{tab:tok}. Results suggest that following the word-piece tokenizer provides superior performance to the character-based alternative. This result is being reflected across all the evaluated metrics, however similarly to the results in Table~\ref{tab:mainres}, we observe the larger gap when considering subjective content preservation study.

\begin{table}[t!]
  \caption{Results for LM trained with chars and word-piece text tokenizers. We report WER, CER, and a Human study for content preservation. We report mean and standard deviations for the human study. \label{tab:tok} }  
  \centering
  \begin{tabular}{l|cc|c}
    \toprule
    & \multicolumn{2}{c|}{\bf Objective} & \multicolumn{1}{c}{\bf Human Study} \\   
    \midrule
    \textbf{Tokenizer} & \textbf{WER} & \textbf{CER} & \textbf{Content}\\
    \midrule    
    Chars       & 0.20    & 0.17 & 2.6 ($\pm 0.81$) \\
    Word-Piece  &0.18     & 0.07 & 3.8 ($\pm 0.75$) \\
    \bottomrule
  \end{tabular}  
  \vspace{-0.2cm}
\end{table}
\section{Conclusion}
\label{sec:con}
In this work, we demonstrate how language models that operate over discrete speech tokens can act as Diacritics-free Hebrew TTS systems. Due to their naturally contextualized manner, language models can better handle ambiguous pronunciations obtained in the absence of diacritics. We empirically show that following the language modeling approach, trained at scale using weakly transcribed data, yields superior performance to non-contextualized, traditional TTS systems when considering context preservation, naturalness, and similarity to the speaker in the generated samples.

\newpara{Limitations.} As the our method is based on auto-regressive LM its inference time is relatively long compared to other TTS systems. Moreover, due to its auto-regressive nature, the duration of the generated speech is determined by the model outputs an end-of-sequence token. Additionally, the model can skip words or invent new ones that did not appear in the text prompt. Although we did not observe such behavior significantly affecting model performance, this lack of controllability imposes another limitation when following the LM approach. 

\newpara{Future work.} To advance research in the field, more benchmarks are needed. For future work, we aim to tackle this task by constructing high-quality, large-scale speech data, directly dedicated for synthesis purposes. 

\newpara{Acknowledgements} This research work was supported by the Israel Innovation Authority, grant number 78563.

\newpage
\bibliographystyle{IEEEtran}
\bibliography{mybib}

\begin{thebibliography}{10}
\providecommand{\url}[1]{#1}
\csname url@samestyle\endcsname
\providecommand{\newblock}{\relax}
\providecommand{\bibinfo}[2]{#2}
\providecommand{\BIBentrySTDinterwordspacing}{\spaceskip=0pt\relax}
\providecommand{\BIBentryALTinterwordstretchfactor}{4}
\providecommand{\BIBentryALTinterwordspacing}{\spaceskip=\fontdimen2\font plus
\BIBentryALTinterwordstretchfactor\fontdimen3\font minus \fontdimen4\font\relax}
\providecommand{\BIBforeignlanguage}[2]{{%
\expandafter\ifx\csname l@#1\endcsname\relax
\typeout{** WARNING: IEEEtran.bst: No hyphenation pattern has been}%
\typeout{** loaded for the language `#1'. Using the pattern for}%
\typeout{** the default language instead.}%
\else
\language=\csname l@#1\endcsname
\fi
#2}}
\providecommand{\BIBdecl}{\relax}
\BIBdecl

\bibitem{campbell2008ethnologue}
L.~Campbell, ``Ethnologue: Languages of the world,'' 2008.

\bibitem{radford2022robust}
A.~Radford, J.~W. Kim, T.~Xu, G.~Brockman, C.~McLeavey, and I.~Sutskever, ``Robust speech recognition via large-scale weak supervision,'' 2022.

\bibitem{pratap2023scaling}
V.~Pratap, A.~Tjandra, B.~Shi, P.~Tomasello, A.~Babu, S.~Kundu, A.~Elkahky, Z.~Ni, A.~Vyas, M.~Fazel-Zarandi, A.~Baevski, Y.~Adi, X.~Zhang, W.-N. Hsu, A.~Conneau, and M.~Auli, ``Scaling speech technology to 1,000+ languages,'' 2023.

\bibitem{izre2001designing}
S.~Izre'el, B.~Hary, and G.~Rahav, ``Designing cosih: the corpus of spoken israeli hebrew,'' \emph{International Journal of Corpus Linguistics}, vol.~6, no.~2, pp. 171--197, 2001.

\bibitem{azogui2016open}
J.~Azogui, A.~Lerner, and V.~Silber-Varod, ``The open university of israel map task corpus (matacop),'' 2016.

\bibitem{marmorstein2022huji}
M.~Marmorstein and N.~Matalon, ``The huji corpus of spoken hebrew: An interaction-oriented design of a corpus,'' 2022.

\bibitem{sharoni2023saspeech}
O.~Sharoni, R.~Shenberg, and E.~Cooper, ``Saspeech: A hebrew single speaker dataset for text to speech and voice conversion,'' in \emph{Proc. Interspeech}, 2023.

\bibitem{wang2023neural}
C.~Wang, S.~Chen, Y.~Wu, Z.~Zhang, L.~Zhou, S.~Liu, Z.~Chen, Y.~Liu, H.~Wang, J.~Li \emph{et~al.}, ``Neural codec language models are zero-shot text to speech synthesizers,'' \emph{arXiv preprint arXiv:2301.02111}, 2023.

\bibitem{marmor2023ivrit}
Y.~Marmor, K.~Misgav, and Y.~Lifshitz, ``ivrit. ai: A comprehensive dataset of hebrew speech for ai research and development,'' \emph{arXiv preprint arXiv:2307.08720}, 2023.

\bibitem{turetzky2024hebdb}
A.~Turetzky, O.~Tal, Y.~Segal-Feldman, Y.~Dissen, E.~Zeldes, A.~Roth, E.~Cohen, Y.~Shrem, B.~R. Chernyak, O.~Seleznova \emph{et~al.}, ``Hebdb: a weakly supervised dataset for hebrew speech processing,'' \emph{arXiv preprint arXiv:2407.07566}, 2024.

\bibitem{Mehta_2023}
\BIBentryALTinterwordspacing
S.~Mehta, A.~Kirkland, H.~Lameris, J.~Beskow, E.~Szekely, and G.~E. Henter, ``Overflow: Putting flows on top of neural transducers for better tts,'' Aug. 2023. [Online]. Available: \url{http://dx.doi.org/10.21437/Interspeech.2023-1996}
\BIBentrySTDinterwordspacing

\bibitem{kong2020hifigan}
J.~Kong, J.~Kim, and J.~Bae, ``Hifi-gan: Generative adversarial networks for efficient and high fidelity speech synthesis,'' \emph{Advances in neural information processing systems}, vol.~33, pp. 17\,022--17\,033, 2020.

\bibitem{défossez2022high}
A.~D{\'e}fossez, J.~Copet, G.~Synnaeve, and Y.~Adi, ``High fidelity neural audio compression,'' \emph{arXiv preprint arXiv:2210.13438}, 2022.

\bibitem{zeghidour2021soundstream}
N.~Zeghidour, A.~Luebs, A.~Omran, J.~Skoglund, and M.~Tagliasacchi, ``Soundstream: An end-to-end neural audio codec,'' \emph{IEEE/ACM Transactions on Audio, Speech, and Language Processing}, vol.~30, pp. 495--507, 2021.

\bibitem{copet2024simple}
J.~Copet, F.~Kreuk, I.~Gat, T.~Remez, D.~Kant, G.~Synnaeve, Y.~Adi, and A.~D{\'e}fossez, ``Simple and controllable music generation,'' \emph{Advances in Neural Information Processing Systems}, vol.~36, 2024.

\bibitem{kreuk2023audiogen}
F.~Kreuk, G.~Synnaeve, A.~Polyak, U.~Singer, A.~D{\'e}fossez, J.~Copet, D.~Parikh, Y.~Taigman, and Y.~Adi, ``Audiogen: Textually guided audio generation,'' \emph{arXiv preprint arXiv:2209.15352}, 2022.

\bibitem{borsos2023audiolm}
Z.~Borsos, R.~Marinier, D.~Vincent, E.~Kharitonov, O.~Pietquin, M.~Sharifi, D.~Roblek, O.~Teboul, D.~Grangier, M.~Tagliasacchi \emph{et~al.}, ``Audiolm: a language modeling approach to audio generation,'' \emph{IEEE/ACM Transactions on Audio, Speech, and Language Processing}, 2023.

\bibitem{sheffer2023hear}
R.~Sheffer and Y.~Adi, ``I hear your true colors: Image guided audio generation,'' in \emph{ICASSP 2023-2023 IEEE International Conference on Acoustics, Speech and Signal Processing (ICASSP)}.\hskip 1em plus 0.5em minus 0.4em\relax IEEE, 2023, pp. 1--5.

\bibitem{kharitonov2023speak}
E.~Kharitonov, D.~Vincent, Z.~Borsos, R.~Marinier, S.~Girgin, O.~Pietquin, M.~Sharifi, M.~Tagliasacchi, and N.~Zeghidour, ``Speak, read and prompt: High-fidelity text-to-speech with minimal supervision,'' \emph{Transactions of the Association for Computational Linguistics}, vol.~11, pp. 1703--1718, 2023.

\bibitem{lyth2024natural}
D.~Lyth and S.~King, ``Natural language guidance of high-fidelity text-to-speech with synthetic annotations,'' \emph{arXiv preprint arXiv:2402.01912}, 2024.

\bibitem{tan2021survey}
X.~Tan, T.~Qin, F.~Soong, and T.-Y. Liu, ``A survey on neural speech synthesis,'' \emph{arXiv preprint arXiv:2106.15561}, 2021.

\bibitem{devlin2019bert}
J.~Devlin, M.-W. Chang, K.~Lee, and K.~Toutanova, ``Bert: Pre-training of deep bidirectional transformers for language understanding,'' 2019.

\bibitem{alephBert2021}
A.~Seker, E.~Bandel, D.~Bareket, I.~Brusilovsky, R.~S. Greenfeld, and R.~Tsarfaty, ``Alephbert: A hebrew large pre-trained language model to start-off your hebrew nlp application with,'' \emph{arXiv preprint arXiv:2104.04052}, 2021.

\bibitem{wu2016google}
Y.~Wu, M.~Schuster, Z.~Chen, Q.~V. Le, M.~Norouzi, W.~Macherey, M.~Krikun, Y.~Cao, Q.~Gao, K.~Macherey \emph{et~al.}, ``Google's neural machine translation system: Bridging the gap between human and machine translation,'' \emph{arXiv preprint arXiv:1609.08144}, 2016.

\bibitem{schuster2012japanese}
M.~Schuster and K.~Nakajima, ``Japanese and korean voice search,'' in \emph{2012 IEEE international conference on acoustics, speech and signal processing (ICASSP)}.\hskip 1em plus 0.5em minus 0.4em\relax IEEE, 2012, pp. 5149--5152.

\bibitem{Gage1994ANA}
\BIBentryALTinterwordspacing
P.~Gage, ``A new algorithm for data compression,'' \emph{The C Users Journal archive}, vol.~12, pp. 23--38, 1994. [Online]. Available: \url{https://api.semanticscholar.org/CorpusID:59804030}
\BIBentrySTDinterwordspacing

\bibitem{kharitonov2022text}
E.~Kharitonov, A.~Lee, A.~Polyak, Y.~Adi, J.~Copet, K.~Lakhotia, T.-A. Nguyen, M.~Rivi{\`e}re, A.~Mohamed, E.~Dupoux \emph{et~al.}, ``Text-free prosody-aware generative spoken language modeling,'' in \emph{ACL 2022-Association for Computational Linguistics}, vol.~1.\hskip 1em plus 0.5em minus 0.4em\relax MIT Press, 2022, pp. 8666--8681.

\bibitem{silero_vad}
S.~Team, ``Silero vad: pre-trained enterprise-grade voice activity detector (vad), number detector and language classifier,'' 2021.

\bibitem{yao2024zipformer}
\BIBentryALTinterwordspacing
Z.~Yao, L.~Guo, X.~Yang, W.~Kang, F.~Kuang, Y.~Yang, Z.~Jin, L.~Lin, and D.~Povey, ``Zipformer: A faster and better encoder for automatic speech recognition,'' in \emph{The Twelfth International Conference on Learning Representations}, 2024. [Online]. Available: \url{https://openreview.net/forum?id=9WD9KwssyT}
\BIBentrySTDinterwordspacing

\bibitem{sanroman2023fromdi}
R.~San~Roman, Y.~Adi, A.~Deleforge, R.~Serizel, G.~Synnaeve, and A.~Défossez, ``From discrete tokens to high-fidelity audio using multi-band diffusion,'' \emph{arXiv preprint arXiv:}, 2023.

\bibitem{Chen_2022}
\BIBentryALTinterwordspacing
S.~Chen, C.~Wang, Z.~Chen, Y.~Wu, S.~Liu, Z.~Chen, J.~Li, N.~Kanda, T.~Yoshioka, X.~Xiao, J.~Wu, L.~Zhou, S.~Ren, Y.~Qian, Y.~Qian, J.~Wu, M.~Zeng, X.~Yu, and F.~Wei, ``Wavlm: Large-scale self-supervised pre-training for full stack speech processing,'' \emph{IEEE Journal of Selected Topics in Signal Processing}, vol.~16, no.~6, p. 1505–1518, Oct. 2022. [Online]. Available: \url{http://dx.doi.org/10.1109/JSTSP.2022.3188113}
\BIBentrySTDinterwordspacing

\bibitem{polyak2021speech}
A.~Polyak, Y.~Adi, J.~Copet, E.~Kharitonov, K.~Lakhotia, W.-N. Hsu, A.~Mohamed, and E.~Dupoux, ``Speech resynthesis from discrete disentangled self-supervised representations,'' \emph{arXiv preprint arXiv:2104.00355}, 2021.

\bibitem{wang2021fairseq}
C.~Wang, W.-N. Hsu, Y.~Adi, A.~Polyak, A.~Lee, P.-J. Chen, J.~Gu, and J.~Pino, ``fairseq s\^{} 2: A scalable and integrable speech synthesis toolkit,'' \emph{arXiv preprint arXiv:2109.06912}, 2021.

\bibitem{baevski2020wav2vec}
A.~Baevski, Y.~Zhou, A.~Mohamed, and M.~Auli, ``wav2vec 2.0: A framework for self-supervised learning of speech representations,'' \emph{Advances in neural information processing systems}, vol.~33, pp. 12\,449--12\,460, 2020.

\bibitem{minaee2024large}
S.~Minaee, T.~Mikolov, N.~Nikzad, M.~Chenaghlu, R.~Socher, X.~Amatriain, and J.~Gao, ``Large language models: A survey,'' \emph{arXiv preprint arXiv:2402.06196}, 2024.

\end{thebibliography}

\end{document}